\documentclass[journal]{IEEEtran}

\usepackage{amsmath}
\usepackage{amssymb}
\usepackage{graphicx}
\usepackage{multirow}
\usepackage{amsfonts}
\usepackage{dsfont}
\usepackage{mathrsfs,amssymb}
\usepackage[english]{babel}
\usepackage[T1]{fontenc}
\usepackage{amsthm}
\usepackage{bbding}
\usepackage{epsfig}

\usepackage{color}

\usepackage{algorithm}
\usepackage{algorithmic}

\usepackage[font=small]{caption}
\usepackage{multicol}
\usepackage{balance}
\usepackage{cite}
\usepackage{subfigure}

 \usepackage{footmisc}

\usepackage{bm}

\begin{document}
\begingroup

\title{Ternary and Binary Quantization for Improved  Classification}

\author{Weizhi Lu, Mingrui Chen, Kai Guo and Weiyu Li

\thanks{W. Lu,  M. Chen and  K. Guo are with the School of Control Science and Engineering, Shandong University, Jinan, China. E-mail:  wzlu@sdu.edu.cn}
\thanks{W. Li is with the  Zhongtai Securities Institute for Financial Studies, Shandong University, Jinan, China. E-mail: liweiyu@sdu.edu.cn}

}

\maketitle




\begin{abstract}

Dimension reduction and data quantization are two important methods for reducing data complexity. In the paper, we study the methodology of first reducing data dimension by random projection and then quantizing the projections to ternary or binary codes, which has been widely applied in classification. Usually, the  quantization will seriously degrade the accuracy of classification due to high quantization errors. Interestingly, however, we observe that  the quantization could provide comparable and often  superior accuracy, as the data to be  quantized are  sparse features  generated with common filters.   Furthermore,  this quantization property could be maintained in the random projections of sparse features, if both the features and random projection matrices are sufficiently sparse.  By conducting extensive experiments, we validate and analyze this intriguing property.

\end{abstract}

\begin{IEEEkeywords}
 ternary quantization, binary quantization, sparse features, random projection, object classification, deep learning
\end{IEEEkeywords}

\IEEEpeerreviewmaketitle

\theoremstyle{definition} \newtheorem{theorem}{Theorem}[]
\theoremstyle{definition} \newtheorem{lemma}[theorem]{Lemma}
\theoremstyle{definition} \newtheorem{definition}[theorem]{Definition}
\theoremstyle{definition} \newtheorem{property}[theorem]{Property}
\theoremstyle{definition} \newtheorem{corollary}[theorem]{Corollary}

\renewcommand{\thefootnote}{\fnsymbol{footnote}}

\section{Introduction}

Large-scale  classification poses great challenges to data storage and computation. To alleviate the problem, a general solution is to reduce data complexity by dimension reduction and quantization. In the paper, we study the methodology of first reducing data dimension by random projection and then quantizing the projections to \{0,1\}-binary or $\{0,\pm1\}$-ternary codes.  Random projection is implemented by  multiplying the data with  random matrices \cite{Bingham01,Fradkin03,Fern03Random}, and   quantization is  realized by zeroing out the elements of small magnitude and unifying  the  elements of large magnitude.

The extreme quantization to binary or ternary codes has  been widely applied in  large-scale retrieval \cite{indyk1998approximate,charikar2002similarity,li2012quantized,li2014coding,li2016quantized,  liu2021ternary,chen2021deep}, where the retrieval accuracy usually exhibits obvious degradation due to high quantization errors.  In the paper, however, we demonstrate that the quantization could achieve comparable and even higher accuracy for common sparse features, such as the ones generated by discrete wavelet transform (DWT) \cite{Mallat2009bk},  discrete cosine transom (DCT)  \cite{Gonzalez06}, and convolutional neural networks (CNN) \cite{sharif2014cnn,donahue2014decaf}. The performance improvement caused by  quantization, simply called quantization gain, could be explained in terms of feature selection. It is noteworthy that the sparse features mentioned above are mainly generated in frequency and/or spatial domains. The small feature elements are usually of high frequencies, mainly caused by noise and edge gradients. The removing of them will help  compact intra-class distances \cite{Zarka2020Deep}, thus improving the classification accuracy. Empirically,  the discrimination between objects  is mainly determined by the distribution of large feature elements over different frequency and/or spacial components,  while insensitive to the  energy of each component. For this reason, the magnitude unification of large elements will not cause serious accuracy degradation, and oftentimes it could even raise the  accuracy thanks to suppressing underlying outliers.



Moreover, it is observed that the quantization gain could  be obtained in the random projections of sparse features,   as both the data features and random matrices are sufficiently sparse, such that the projections are sparse. The sparse projections approximately inherit the sparse structure of the original sparse feature and thus can provide similar quantization gains.   To generate sparse projections,  we suggest to use extremely  sparse $\{0,\pm 1\}$-matrices \cite{Achlioptas03, Li06} for random projection, such as the ones with only one nonzero element per column, which  could provide  desired optimization gains.  In contrast, another popular random projection matrices, Gaussian matrices \cite{dasgupta2003elementary} can hardly obtain such gains, due to always generating Gaussian-like dense projections.

Here we mainly evaluate the classification performance with the exemplar-based classifiers \cite{nosofsky1991tests, peters2003human, ashby2017neural, bowman2020tracking}, also known as the instance \cite{cost1993weighted} or nearest neighbor-based classifiers \cite{Cover67NNC}, which   have been widely recognized as the most biologically-plausible cognitive method \cite{ashby2017neural}. The method categorizes a novel object by comparing its similarity to the exemplars previously stored by class, and could achieve the Bayes-optimal performance, as the exemplars are sampled densely \cite{Cover67NNC}. There are two major ways  to measure the similarity  to each class. One is to evaluate the distribution of a few most similar exemplars across all classes, such as the known $k$ nearest neighbors ($k$NN) classifier \cite{peterson2009k},  and the other is to calculate the  Euclidean distance to the subspace of each class, such as the  local subspace classifier (LSC) \cite{laaksonen1997local} and its  variants \cite{naseem2010linear,veenman2005nearest,cevikalp2007local,lu2010nearest,liu2011k,wang2017consistency}. It is easy to see that $k$NN has  much lower complexity than LSC, but often suffers inferior performance.

To achieve a balance between them,    we propose a simplified variant of LSC, named $k$ nearest exemplar-based classifier ($k$NEC),  which measures the similarity to each class by simply summing the distances to  $k$ nearest exemplars in the class.   Despite the simplicity, $k$NEC often could achieve comparable or  better performance than LSC. In fact, the distance summing method adopted in $k$NEC has been early proposed in \cite{nosofsky1988exemplar}, which however considers the sum of all rather than a few exemplars. The slight modification of $k$NEC could provide significant performance gains, mainly because in practice a query object is often similar to a few rather than all exemplars in its category. The results mentioned above are validated by conducting extensive classification experiments on three benchmark image datasets, including YaleB \cite{Lee05}, Cifar10 \cite{Krizhevsky09Cifar10} and ImageNet \cite{ImageNet09}.

  \section{Method}
In this section, we describe the pipeline of sparse features generation, projection, quantization and classification.

  \subsection{Sparse feature generation}
  \label{secsub:SFG}

To obtain relatively good and meaningful  classification performance, we suggest to select sparse  feature generators in terms of data complexity. For simple datasets, like YaleB, we can simply use linear filters, such as DWT and DCT;  and otherwise, we need more sophisticated filters, like convolutional neural networks \cite{sharif2014cnn,donahue2014decaf}, for more complex datasets, such as Cifar10 and ImageNet.  These features capture the characteristics of original images in  frequency and/or spacial domains, which as shown in Figure \ref{fig:feature},  tend to present  heavy-tailed symmetric distributions \cite{reininger1983distributions,simoncelli1999modeling,garriga2018deep}. Usually, the distribution with sharper peak and longer tails means a more sparse structure.  By the feature distributions illustrated in Figure \ref{fig:feature},  we  can rank the feature sparsity of three datasets in descending order: YaleB, ImageNet and Cifar10. Interestingly, as will be seen later, the data feature of higher sparsity seems more likely to induce quantization gains.


  \subsection{Sparse matrices-based random projection}

For a sparse feature vector  $\bm{x}\in \mathbb{R}^n$,  its random projection is derived by $\bm{y}=\bm{R}\bm{x}$, where $\bm{R}\in \mathbb{R}^{m\times n}$ denotes a random matrix, $m\ll n$. As stated before, like sparse features $x$, sparse projections $y$ tend to provide quantization gains. To generate sparse projections  $\bm{y}$, by  \cite[Sec.2.3]{kotz2012laplace}, we propose to employ  sparse ternary matrices $\bm{R}\in \{0,\pm 1\}^{m\times n}$.  Empirically, the quantization gain tends to first increase and then decrease, as the  matrix becomes more and more sparse.  The latter decreasing trend is incurred by  the increasing loss of important features. Empirically, we could obtain a good quantization performance when assigning only one nonzero element in each matrix column. For comparison, we also test Gaussian matrices-based random projections,  which always have dense projections  and can hardly provide quantization gains.

  \subsection{Ternary and binary quantization}

Given a sparse feature $x$ which has been centralized by mean subtraction, we propose to generate the ternary codes  by  $z_i=sign(x_i)$, if $|x_i|>\tau$ and otherwise, $z_i=0$.  Then, the  binary codes are generated by zeroing out all  negative entries in the ternary codes.  With the same methods, we could generate the codes for  random projections $y$.  To evaluate the influence of code sparsity, we define by $k/n$ the sparsity ratio of $n$-dimensional ternary code, where $k$ counts the number of nonzeros in the code. Accordingly, the sparsity ratio of binary codes should be about $k/2n$.

  \subsection{Exemplar-based classification}
  
  The classification is realized with the proposed $k$NEC, which as stated before, measures the similarity of a query object to each class with the sum of the correlations to $k$-nearest neighbors in the class. Empirically, $k$NEC performs consistently better  than $k$NN, and performs comparably or even better than LSC. The reason is as follows.  Compared to $k$NN that considers only $k$ nearest exemplars among all classes, $k$NEC involves  more exemplars, i.e. $k$ exemplars each class, thus  more robust to noise and outlier.  As for LSC, it measures the similarity  to each class with the Euclidean distance to the subspace spanned by $k$ nearest exemplars in the class. The distance needs to be computed with the minimum square method, while the method inclines to enforcing the  $k$ exemplars to contribute equally in representing the query object. However, the constraint is not  reasonable, if the similarity of the query object to  $k$  exemplars  differs widely. In this case, a survey on the distance to each individual exemplar, as done in $k$NEC, should be more reasonable.

 \begin{figure}[t]
\centering

\subfigure[YaleB]{
\begin{minipage}[t]{0.3\linewidth}
\centering
\includegraphics[height=2.2cm,width=2.85cm]{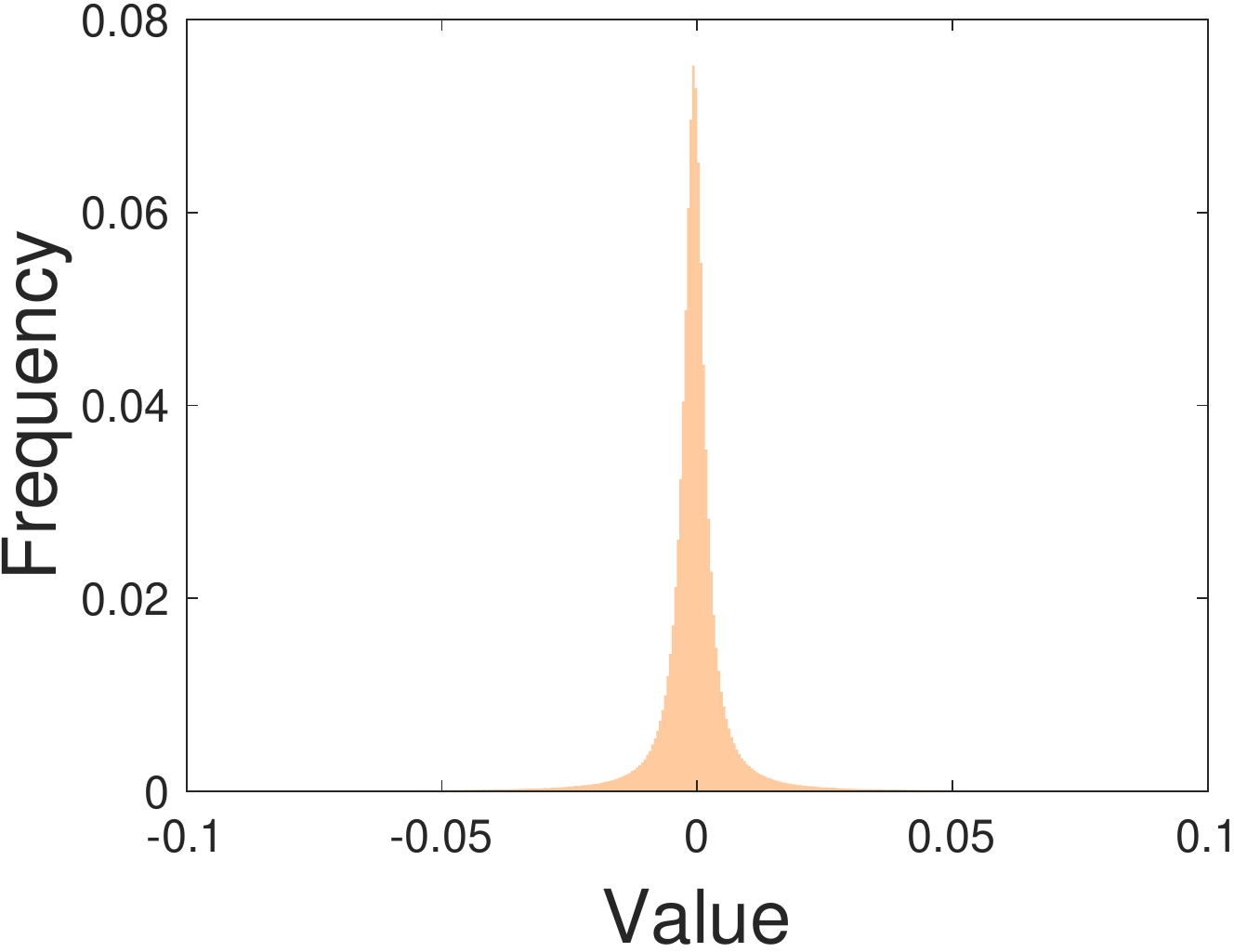}
\end{minipage}
}%
\subfigure[Cifar10]{
\begin{minipage}[t]{0.3\linewidth}
\centering
\includegraphics[height=2.2cm,width=2.85cm]{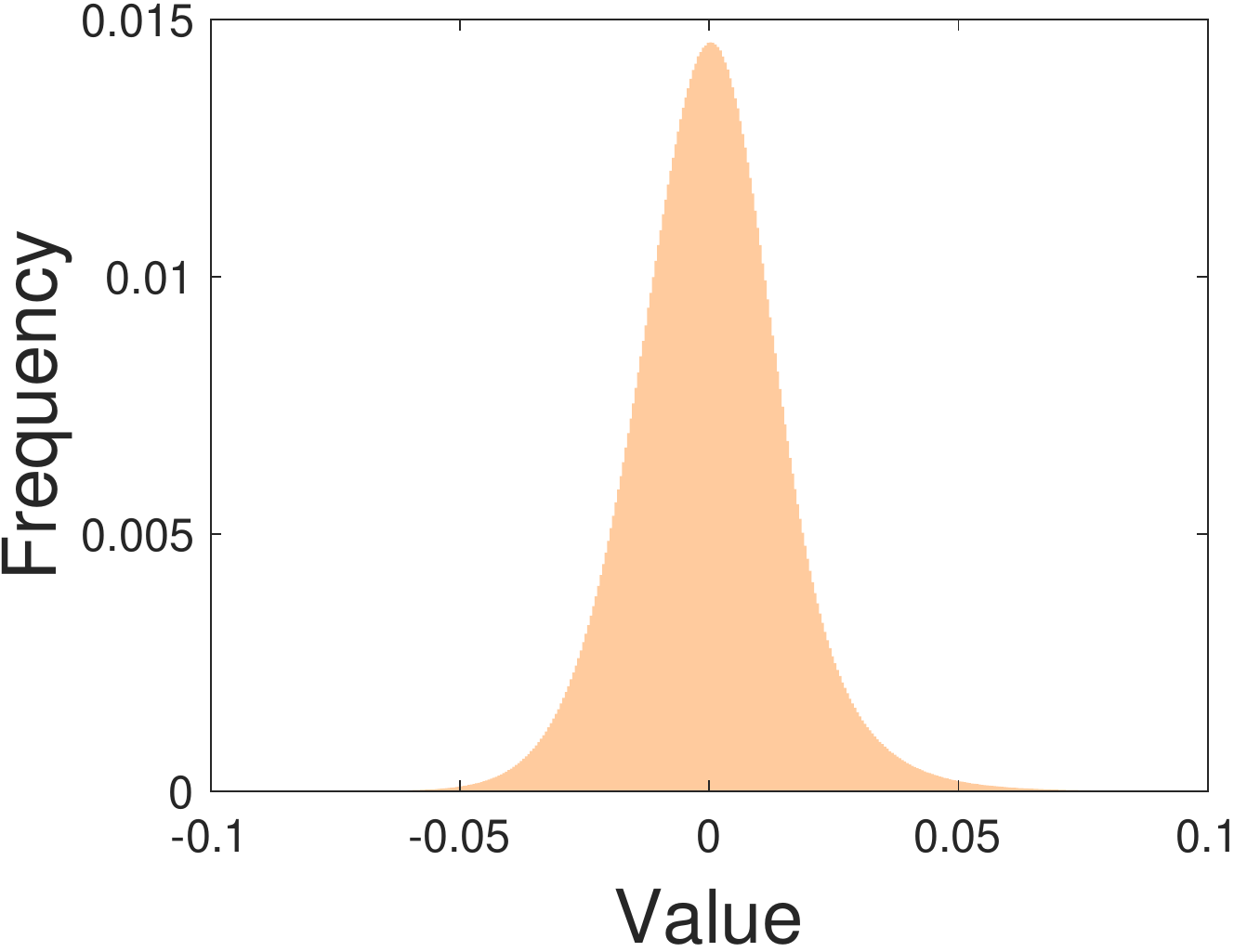}
\end{minipage}
}%
\subfigure[ImageNet]{
\begin{minipage}[t]{0.3\linewidth}
\centering
\includegraphics[height=2.2cm,width=2.85cm]{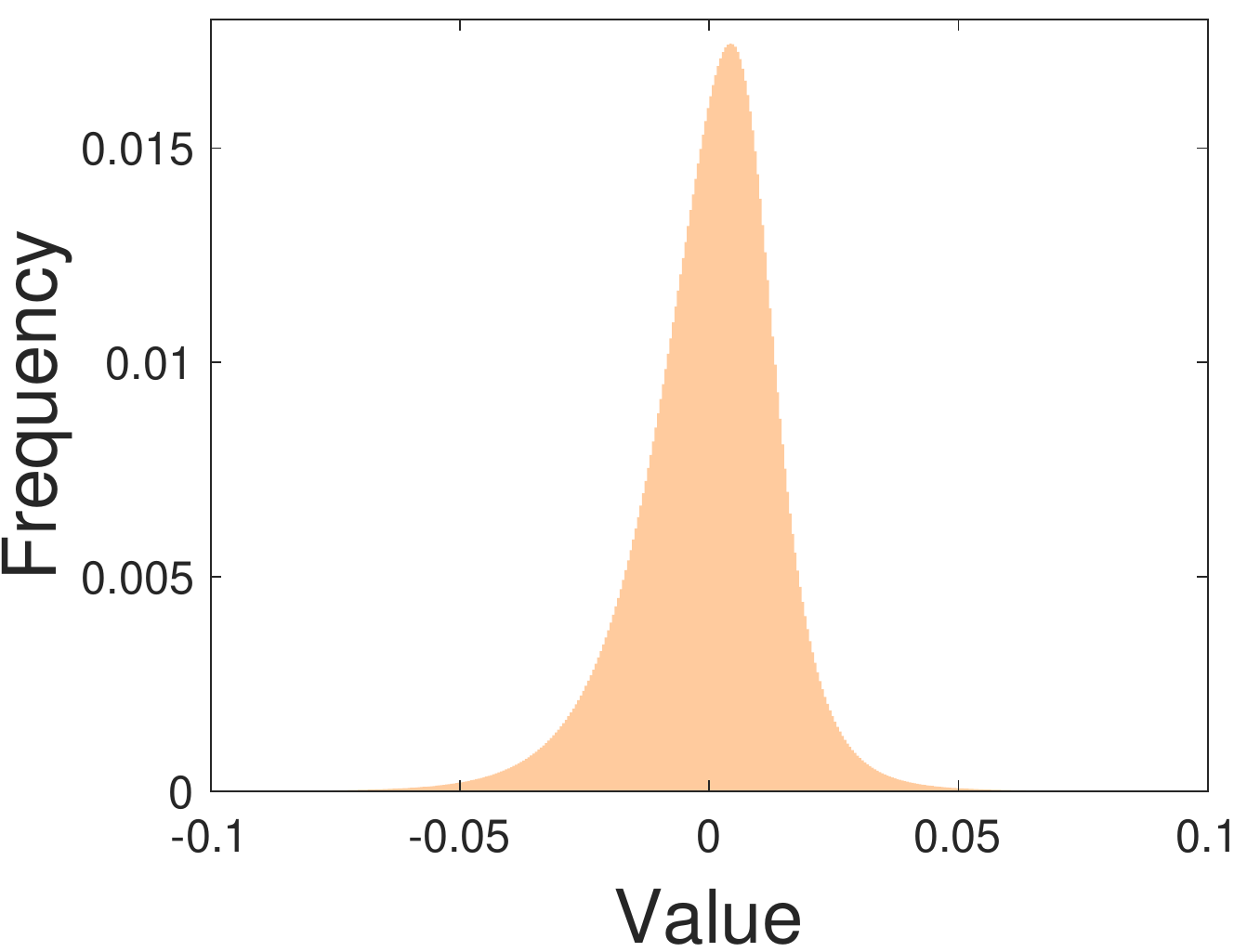}
\end{minipage}
}%

\caption{The  frequency histograms of the sparse features of YaleB, Cifar10, and ImageNet, which are generated respectively by DCT, AlexNet Conv5 and VGG16 Conv$5\_3$.}
\label{fig:feature}
\end{figure}

 \section{Experiments}
\label{sec_experiments}

\begin{table*}[htp]
	\centering
	{
   \begin{tabular}{c|c|c|c c c c c c c c c c}\hline
		     &Sparsity ratio  &$k/n$ & 0.1& 0.2 &0.3&0.4&0.5&0.6&0.7&0.8&0.9&1.0\\ \hline

\multirow{12}{*}{YaleB}
&\multirow{3}{*}{$k$NEC}
& RC & 95.63& 96.96& 97.44& 97.52& 97.61& 97.73& 97.76& 97.78& 97.78& 97.76\\
&& TC & \textbf{99.77}& \textbf{99.93}& \textbf{99.96}& \textbf{99.99}& \textbf{100.00}& \textbf{100.00}& \textbf{99.98}& \textbf{100.00}& \textbf{99.96}& \textbf{99.96}\\
&& BC & \underline{99.36}& \underline{99.78}& \underline{99.74}& \underline{99.83}& \underline{99.87}& \underline{99.94}& \underline{99.96}& \underline{99.98}& \underline{99.94}& \underline{99.96}\\ \cline{2-13}
&\multirow{3}{*}{$k$NN}
& RC & 93.41& 94.92& 95.37& 95.71& 95.69& 95.74& 95.79& 95.81& 95.79& 95.79\\
&& TC & \textbf{98.90}& \textbf{99.56}& \textbf{99.77}& \textbf{99.82}& \textbf{99.94}& \textbf{99.94}& \textbf{99.86}& \underline{99.87}& \textbf{99.96}& \textbf{99.86}\\
&& BC & \underline{97.69}& \underline{99.12}& \underline{99.23}& \underline{99.60}& \underline{99.74}& \underline{99.75}& \underline{99.78}& \textbf{99.92}& \underline{99.86}& \underline{99.86}\\ \cline{2-13}
&\multirow{3}{*}{LSC}
& RC & 98.89& 99.15& 99.18& 99.21& 99.24& 99.24& 99.20& 99.20& 99.19& 99.20\\
&& TC & \textbf{99.81}& \textbf{99.96}& \textbf{99.96}& \textbf{100.00}& \textbf{100.00}& \textbf{100.00}& \textbf{100.00}& \textbf{100.00}& \textbf{99.96}& \textbf{99.98}\\
&& BC & \underline{99.58}& \underline{99.88}& \underline{99.91}& \underline{99.95}& \underline{99.97}& \underline{99.95}& \underline{99.94}& \underline{99.95}& \underline{99.91}& \underline{99.94}\\ \cline{2-13}
&\multirow{3}{*}{SVM}
& RC & 85.03& 90.66& 92.12& 92.77& 93.24& 93.41& 93.64& 93.68& 93.67& 93.67\\
&& TC & \textbf{99.18}& \textbf{99.82}& \textbf{99.93}& \textbf{99.96}& \textbf{99.94}& \textbf{99.94}& \textbf{99.96}& \textbf{99.95}& \textbf{99.96}& \textbf{99.94}\\
&& BC & \underline{98.68}& \underline{99.69}& \underline{99.84}& \underline{99.93}& \underline{99.92}& \underline{99.92}& \underline{99.90}& \underline{99.90}& \underline{99.94}& \underline{99.94}\\ \hline
\multirow{12}{*}{Cifar10}
&\multirow{3}{*}{$k$NEC}
& RC & 72.39& 74.70& 75.46& \textbf{76.16}& \textbf{76.57}& \textbf{76.79}& \textbf{76.61}& \textbf{76.67}& \textbf{76.66}& \textbf{76.71}\\
&& TC & \underline{72.48}& \underline{74.81}& \underline{75.59}& \underline{76.02}& 75.93& 75.76& 75.80& \underline{75.42}& \underline{75.58}& 75.08\\
&& BC & \textbf{72.99}& \textbf{75.15}& \textbf{75.82}& 75.96& \underline{76.16}& \underline{76.06}& \underline{75.82}& 75.31& 75.38& \underline{75.10}\\ \cline{2-13}
&\multirow{3}{*}{$k$NN}
& RC & \textbf{69.21}& \textbf{71.56}& 72.36& \underline{72.88}& \textbf{73.69}& \textbf{73.69}& \textbf{73.83}& \textbf{73.83}& \textbf{73.86}& \textbf{73.78}\\
&& TC & 68.24& \underline{71.33}& \textbf{72.73}& 72.71& 72.66& 72.57& \underline{72.89}& 72.43& \underline{72.53}& \underline{71.80}\\
&& BC & \underline{68.53}& 71.08& \underline{72.66}& \textbf{73.18}& \underline{73.25}& \underline{72.94}& 72.84& \underline{72.53}& 72.16& 71.80\\ \cline{2-13}
&\multirow{3}{*}{LSC}
& RC & \underline{74.42}& 76.46& \underline{77.28}& \underline{77.74}& \underline{78.05}& \textbf{77.98}& \textbf{78.18}& \textbf{78.12}& \textbf{78.20}& \textbf{78.12}\\
&& TC & \textbf{74.70}& \underline{76.58}& \textbf{77.50}& \textbf{77.87}& \textbf{78.06}& \underline{77.65}& \underline{77.88}& \underline{77.65}& \underline{77.31}& \underline{77.35}\\
&& BC & 74.26& \textbf{76.90}& 77.26& 77.01& 76.94& 76.79& 77.02& 76.13& 76.09& 75.92\\ \cline{2-13}
&\multirow{3}{*}{SVM}
& RC & \textbf{77.65}& \textbf{79.24}& \textbf{79.77}& \textbf{80.00}& \textbf{80.09}& \textbf{80.37}& \textbf{80.58}& \textbf{80.52}& \textbf{80.56}& \textbf{80.66}\\
&& TC & \underline{72.28}& \underline{74.74}& \underline{74.92}& \underline{75.41}& \underline{75.70}& \underline{76.10}& \underline{76.42}& \underline{75.61}& \underline{75.09}& \underline{74.44}\\
&& BC & 71.94& 74.24& 74.37& 74.95& 75.15& 74.52& 74.36& 73.54& 73.90& 74.36\\ \hline
\multirow{6}{*}{ImageNet}
&\multirow{3}{*}{$k$NEC}
& RC & 39.05& 40.73& \underline{41.79}& \textbf{42.32}& \textbf{42.61}& \textbf{42.64}& \textbf{42.71}& \textbf{42.70}& \textbf{42.50}& \textbf{42.31}\\
&& TC & \underline{42.26}& \underline{42.20}& 41.43& 40.20& \underline{38.84}& \underline{37.54}& \underline{36.19}& \underline{35.24}& 34.28& \underline{33.74}\\
&& BC & \textbf{48.28}& \textbf{45.25}& \textbf{42.61}& \underline{40.55}& 38.67& 37.13& 36.01& 35.19& \underline{34.39}& 33.74\\ \cline{2-13}
&\multirow{3}{*}{$k$NN}
& RC & 36.69& 38.49& \underline{39.26}& \textbf{39.68}& \textbf{39.95}& \textbf{39.97}& \textbf{40.02}& \textbf{39.93}& \textbf{39.78}& \textbf{39.65}\\
&& TC & \underline{39.74}& \underline{39.70}& 38.77& 37.53& \underline{36.16}& \underline{34.64}& \underline{33.52}& 32.33& 31.48& \underline{31.00}\\
&& BC & \textbf{46.07}& \textbf{42.76}& \textbf{40.01}& \underline{37.80}& 35.81& 34.30& 33.27& \underline{32.42}& \underline{31.58}& 31.00\\ \hline

	\end{tabular}
	}
	\caption{Classification accuracy of four classifiers on real-valued codes (RC), ternary codes (TC) and binary codes (BC)  across varying sparsity ratio $k/n$. The codes are generated with the sparse features of three different datasets. The best results are highlighted in bold and the second best are underlined.}
	\label{tab:SF}
\end{table*}

In this section, we  evaluate the quantization performance for  the sparse features of  YaleB \cite{Lee05}, CIFAR10 \cite{Krizhevsky09Cifar10} and ImageNet \cite{ImageNet09}, which are generated respectively by DCT, AlexNet Conv5 and VGG16 Conv$5\_3$ \cite{simonyan2014very}, with vectorized dimensions 32256, 43264 and 100352. To reduce simulation time,  we further downsample the features to 1200, 4327, and 5018 dimensions.  This may cause accuracy degradation but will not influence our comparative studies.  The experimental settings are  briefly introduced as follows. YaleB contains 2414 frontal-face images with size $192\times 168$ over 38 subjects and about 64 images per subject. We randomly select 9/10 of samples for training and the rest for testing. Considering the DCT and DWT features of YaleB present similar performance, we only present the results of DCT due to limited space.  Cifar10  consists of $32\times32$ natural color images in 10 classes, with 50k samples for training and 10k samples for testing. The same data division is adopted in our experiments. ImageNet contains 1000 classes of images with average resolution $469\times387$, which has about 1.2M samples for training, 50k samples for validation, and 100k samples for testing. Here we take the  validation set  for testing since its labels are available.

 For comparison, besides the proposed $k$NEC, we also test three other popular classifiers, $k$NN \cite{peterson2009k}, LSC \cite{laaksonen1997local}, and SVM \cite{cortes1995support}. For $k$NEC, LSC and  $k$NN, we need to previously determine  the number $k$ of the nearest neighbors in each class ($k$NEC, LSC) or among all classes ($k$NN).  Empirically, it suffices to achieve a good performance when  simply setting  $k=3$ or 5. Note that we do not test LSC and SVM for ImageNet, due to prohibitive computation. In the following, we evaluate the quantization performance respectively for sparse features and their random projections.

\begin{table*}[htp]
	\centering
	{
   \begin{tabular}{c|c|c|c c c c c c c c c c}\hline
		     & Sparsity ratio & $k/n$ & 0.1& 0.2 &0.3&0.4&0.5&0.6&0.7&0.8&0.9&1.0\\ \hline
		
\multirow{6}{*}{YaleB}
&\multirow{3}{*}{Sparse Matrix}
& RC & 96.15& 97.02& 97.27& 97.36& 97.45& 97.51& 97.53& 97.54& 97.53& 97.53\\
&& TC & \textbf{99.64}& \textbf{99.89}& \textbf{99.93}& \textbf{99.91}& \textbf{99.94}& \textbf{99.94}& \textbf{99.92}& \textbf{99.93}& \textbf{99.92}& \textbf{99.91}\\
&& BC & \underline{98.50}& \underline{99.47}& \underline{99.65}& \underline{99.69}& \underline{99.64}& \underline{99.73}& \underline{99.70}& \underline{99.74}& \underline{99.75}& \underline{99.74}\\ \cline{2-13}
&\multirow{3}{*}{Gaussian Matrix}
& RC & \textbf{95.85}& \textbf{96.90}& \textbf{97.22}& \textbf{97.43}& \textbf{97.45}& \textbf{97.49}& \textbf{97.53}& \textbf{97.55}& \textbf{97.54}& \textbf{97.53}\\
&& TC & \underline{93.67}& \underline{95.20}& \underline{95.74}& \underline{95.96}& \underline{95.96}& \underline{96.10}& \underline{95.96}& \underline{95.97}& \underline{96.14}& \underline{96.18}\\
&& BC & 88.96& 91.00& 92.10& 92.52& 92.48& 92.46& 92.50& 92.52& 92.68& 92.82\\ \hline
\multirow{6}{*}{Cifar10}
&\multirow{3}{*}{Sparse Matrix}
& RC & \textbf{71.64}& \textbf{74.11}& \textbf{75.07}& \textbf{75.63}& \textbf{75.95}& \textbf{76.12}& \textbf{76.21}& \textbf{76.24}& \textbf{76.28}& \textbf{76.29}\\
&& TC & \underline{70.79}& \underline{73.01}& \underline{74.09}& \underline{74.69}& \underline{74.86}& \underline{75.23}& \underline{75.31}& \underline{75.39}& \underline{75.41}& \underline{75.48}\\
&& BC & 67.25& 70.50& 72.31& 72.32& 72.68& 72.81& 73.20& 73.20& 73.21& 73.25\\ \cline{2-13}
&\multirow{3}{*}{Gaussian Matrix}
& RC & \textbf{70.52}& \textbf{72.97}& \textbf{74.09}& \textbf{74.63}& \textbf{75.05}& \textbf{75.34}& \textbf{75.38}& \textbf{75.46}& \textbf{75.54}& \textbf{75.52}\\
&& TC & \underline{69.26}& \underline{71.77}& \underline{72.99}& \underline{73.72}& \underline{74.11}& \underline{74.35}& \underline{74.55}& \underline{74.59}& \underline{74.64}& \underline{74.58}\\
&& BC & 66.02& 68.61& 69.82& 70.92& 71.56& 72.01& 72.34& 72.39& 72.45& 72.41\\ \hline
\multirow{6}{*}{ImageNet}
&\multirow{3}{*}{Sparse Matrix}
& RC & \underline{38.54}& \underline{40.35}& \textbf{41.28}& \textbf{41.86}& \textbf{42.05}& \textbf{42.08}& \textbf{42.16}& \textbf{42.07}& \textbf{41.94}& \textbf{41.78}\\
&& TC & \textbf{40.70}& \textbf{40.79}& \underline{40.42}& \underline{40.33}& \underline{40.36}& \underline{40.24}& \underline{40.09}& \underline{39.92}& \underline{39.79}& \underline{39.51}\\
&& BC & 36.19& 36.12& 37.18& 37.13& 37.20& 37.27& 37.14& 36.92& 36.68& 36.39\\ \cline{2-13}
&\multirow{3}{*}{Gaussian Matrix}
& RC & \textbf{37.82}& \textbf{39.61}& \textbf{40.76}& \textbf{41.17}& \textbf{41.47}& \textbf{41.51}& \textbf{41.56}& \textbf{41.50}& \textbf{41.40}& \textbf{41.27}\\
&& TC & \underline{37.26}& \underline{39.01}& \underline{39.96}& \underline{40.42}& \underline{40.85}& \underline{40.94}& \underline{40.99}& \underline{41.01}& \underline{40.78}& \underline{40.61}\\
&& BC & 35.18& 36.88& 38.03& 38.41& 38.75& 38.99& 39.09& 39.04& 39.07& 38.80\\ \hline

	\end{tabular}
	}
	\caption{Classification accuracy of $k$NEC on real-valued codes (RC), ternary codes (TC) and binary codes (BC)  across varying sparsity ratio $k/n$. The codes are generated with the random projections of  sparse features over sparse ternary matrices and Gaussian matrices. The projection matrices have compression ratio $m/n=1/2$. The best results are highlighted in bold and the second best are underlined. }
	\label{tab:RP}
\end{table*}

\subsection{Quantization of  sparse features}
In this part, we aim to prove two facts: 1) the two steps involved in our ternary quantization, namely the removing of small elements and the magnitude unification  of large elements, could improve the classification accuracy of sparse features; 2) the proposed $k$NEC could achieve better performance than $k$NN and comparable performance with LSC. To this end, besides  ternary codes (TC) and binary codes (BC), we also consider a  kind of real-valued codes (RC), which has the same sparsity with TC but has the nonzero values un-quantized. This means that  RC will be identical to the original sparse features, when the sparsity ratio $k/n=1$. Note the lower sparsity ratio means higher sparsity.

The results are shown in Table \ref{tab:SF}. It is seen that  in three datasets, the performance of RC  tends to first increase and then decrease with the sparsity ratio $k/n$ decreasing from 1 to 0.1. The initial increasing trend suggests that the removing of small elements indeed could provide performance gains, while the latter decreasing  is incurred by the increasing loss of significant feature elements. Comparing the performance of RC and TC, we can see that TC performs better at low sparsity ratios $k/n\leq 0.3$ both in Cifar10 and ImageNet, and consistently better in YaleB. The performance advantage of TC over RC suggests that the magnitude unification of large elements involved in TC indeed could improve the classification accuracy. Compared to TC, BC discard about half of the features (the negative part) but only suffer from  about 0.5\% accuracy loss in most cases.  The similarity between them  implies that the positive  and negative parts of sparse features may share similar features. This is indirectly confirmed by the symmetric distributions of  sparse features, as illustrated in Figure \ref{fig:feature}.  In ImageNet,  BC even outperform TC at $k/n\leq 0.4$. This may be explained by the fact that the sparse features with mean substraction are not exactly symmetric about zero, and it is imperfect to generate TC using the same threshold $\tau$ in both sides.

As for the classifiers, Table \ref{tab:SF} shows that  $k$NEC consistently outperforms  $k$NN, and performs comparably or better than LSC, with gaps around 1\%.  In YaleB, $k$NEC even outperforms SVM. The superiority  should be attributed to the high similarity between intra-class face samples, which is favorable for exemplar-based classification.  In three datasets, three exemplar-based classifiers all could achieve quantization gains,  while SVM fails in Cifar10. Furthermore, it is noteworthy that $k$NEC and LSC tend to  outperform SVM, when all of them use ternary or binary codes.  This suggests that the quantization may be more matched with  exemplar-based classification than with SVM. The perfect match between sparse codes and  exemplar-based classification may be attributed to its biological plausibility in visual feature extraction \cite{Field96Nature,olshausen1997sparse,van1998independent,hyvarinen2007complex} and recognition \cite{ashby2017neural}.

For the thee datasets, by their capability of inducing optimization gains as shown in Table \ref{tab:SF}, we can rank them in descending order: YaleB, ImageNet and Cifar10. Interestingly, the order  happens to agree with the feature sparsity order  we have ranked with the feature distributions shown in Figure \ref{fig:feature}.  As stated in Section \ref{secsub:SFG}, this implies that the features with higher sparsity, namely the ones with sharper peaks and longer tails, are  more prone to providing quantization gains. 

Finally, it is noteworthy $k$NEC could achieve comparable or even better classification performance than deep networks, if exploiting sparse features from  deeper network layers. For instance, using the output  features of the penultimate, fully-connected layer,  for Cifar10 with classification accuracy 89.02\% on AlexNet,  we obtain the $k$NEC accuracy of 89.66\%, 89.52\% and 89.50\%   respectively for RC, TC and BC all with sparsity ratio $k/n=10\%$; and  for ImageNet with classification accuracy 65.76\% on VGG16, $k$NEC  achieves the accuracy of 65.54\%, 65.17\% and 65.24\% for the above-mentioned three codes with $k/n=30\%$.

\subsection{Quantization of  sparse random projections}
We now move on to prove that the quantization gain could also be obtained for the random projections of sparse features, if both the features and random projection matrices are sufficiently sparse, such that the projections are sparse. For this purpose, we test the sparse features with varying sparsity ratio $k/n$, and implement random projections with a kind of  extremely sparse random matrices $\bm{R}\in\{0,\pm 1\}^{m\times n}$, which has only one nonzero entry per column. For comparison, we also test  Gaussian matrices based random projections, which always have dense distributions.   To obtain relatively stable classification results \cite{Bingham01},  we  obtain an average result in each trial by repeating 5 times random projections.  

The results are provided in Table \ref{tab:RP}. It is seen that sparse matrices achieve quantization  gains with TC both in YaleB ($k/n\leq 1$) and ImageNet ($k/n\leq 0.2$). As to why the gain is not obtained in Cifar10, it may be because its features have relatively weak sparsity nature compared to the features of other two datasets, as illustrated in Figure \ref{fig:feature}. As demonstrated before, the problem should be addressed if we explore deeper networks to generate more abstract and sparse features for Cifar10.  As for Gaussian matrices, it is seen that the best performance is always obtained by RC, rather than by TC or BC. This suggests that Gaussian matrices can hardly offer quantization gains due to holding dense projections.  Furthermore, it is noteworthy that the RC performance of Gaussian matrices  is always inferior to the TC performance of sparse matrices. This encourages us to exploit sparse matrices instead of  Gaussian matrices for quantization, both in terms of  performance and complexity.

\section{Conclusion and Discussion}
For sparse features or their sparse random projections, we have shown that   ternary quantization on them could improve classification performance, and  binary quantization could  provide competitive performance. The intriguing quantization performance can be explained with feature selection: the zero out of small elements helps reduce noise and compact intra-class distances, while the magnitude unification of large elements helps remove underlying outliers. Interestingly,  the  quantization tends to provide better classification performance when combined with  exemplar-based classifiers, than with other types of classifiers, such as SVM. The perfect match between sparse codes and exemplar-based classifiers may be explained by its biological plausibility: it has been found that the human vision system  captures visual features  with Gabor-like sparse binary codes \cite{Field96Nature,olshausen1997sparse,van1998independent,hyvarinen2007complex}, and recognizes them with  exemplar-based cognitive mechanism \cite{ashby2017neural}. Finally, it is noteworthy that in our classification  both  the initial random projection of sparse features  and the following exemplar-based classification are linear projection models. Essentially, the cascade of the two models forms a  three-layer neural network, with ternary/binary parameters/activations. Usually, such kind of  quantization networks performs comparably or even better than its full-precision counterparts \cite{qin2020binary}, while the mechanism behind the intriguing  performance remains unclear. Our work sheds  light on the  problem.

\bibliographystyle{IEEEtran}
\balance
\bibliography{refs_20220317}

\end{document}